\documentclass[10pt,twocolumn,letterpaper]{article}

\usepackage{epsfig}
\usepackage{graphicx}
\usepackage{subfigure}
\usepackage{amsmath}
\usepackage{amssymb}
\usepackage{amsthm}
\usepackage{amsfonts}
\usepackage{mathrsfs}
\usepackage{booktabs}
\usepackage{multirow, eucal}
\usepackage{helvet}
\usepackage{courier}
\usepackage{bm}
\usepackage{bbm}
\usepackage{graphicx}
\usepackage{color}
\usepackage{epstopdf}
\usepackage{wrapfig}
\usepackage{picinpar}
\usepackage{url}

\usepackage[vlined,ruled,linesnumbered]{algorithm2e}
\usepackage{cite}

\usepackage[pagebackref=true,breaklinks=true,letterpaper=true,colorlinks,bookmarks=false]{hyperref}
\usepackage{caption}
\def\diag{\mbox{diag}}
\def\rank{\mbox{rank}}
\def\grad{\mbox{\text{grad}}}
\def\dist{\mbox{dist}}
\def\sgn{\mbox{sgn}}
\def\tr{\mbox{tr}}
\def\etal{{\em et al.\/}\, }
\def\card{{\mbox{Card}}}

\def\0{{\bf 0}}
\def\1{{\bf 1}}

\def\bs{{\bf s}}

\def\bx{{\bf x}}

\def\citep{\cite}
\def\citet{\cite}

\def\diag{\mbox{diag}}
\def\rank{\mbox{rank}}
\def\grad{\mbox{\text{grad}}}
\def\dist{\mbox{dist}}
\def\sgn{\mbox{sgn}}
\def\tr{\mbox{tr}}
\def\etal{{\em et al.\/}\, }
\def\card{{\mbox{Card}}}
\def\st{\mbox{s.t. }}

\newcommand*\xor{\mathbin{\oplus}}

\def\kui{\textcolor{black}}
\def\bohan{\textcolor{black}}

\def\ie{{\it i.e.}}
\def\eg{{\it e.g.}}

\begin{document}

\title{Structured Binary Neural Networks for Accurate Image Classification and Semantic Segmentation}

\author{
    Bohan Zhuang, Chunhua Shen, Mingkui Tan, Lingqiao Liu,  and Ian Reid
}

\maketitle

\begin{abstract}

	In this paper, we propose to train \kui{convolutional neural networks (CNNs) with both binarized weights and
	activations, leading to quantized models  specifically}  for mobile devices with limited  \kui{power capacity and computation resources}.
	Previous works on quantizing CNNs seek to approximate the floating-point information using a set of discrete values, \kui{which we call value approximation}, but typically assume the same architecture as the full-precision networks.

	In this paper, however, we take a novel ``structure approximation'' view for quantization--- it is very likely that a different architecture may be better for best performance. In particular, we propose a ``network decomposition'' strategy, named \textbf{Group-Net}, in which we divide the network into groups.
	In this way, each full-precision group can be effectively reconstructed by aggregating a set of homogeneous binary branches.
	In addition, we learn effective connections among groups to improve the \bohan{representational capability}.
	\bohan{Moreover,} the proposed Group-Net shows strong generalization to other tasks.
    \bohan{For instance,  we extend Group-Net for highly accurate semantic segmentation by embedding rich context into the binary structure.}
	Experiments on both classification and %
	\bohan{semantic segmentation} tasks demonstrate the superior performance of the proposed methods over various popular architectures. In particular, we outperform the previous best binary neural networks in terms of accuracy and major computation savings.

\end{abstract}

\section{Introduction}

\noindent Designing deeper and wider convolutional neural networks has led to significant breakthroughs in many machine learning tasks, such as image classification~\cite{krizhevsky2012imagenet, he2016deep}, object detection~\cite{redmon2016you, ren2015faster} and object segmentation~\cite{long2015fully, chen2018encoder}. However, accurate deep models \kui{often} require billions of FLOPs, which makes it infeasible \kui{for deep models} to run many real-time applications on resource constrained mobile platforms. To solve this problem, many existing works focus on network pruning~\cite{he2017channel, han2015learning}, low-bit quantization~\cite{zhuang2018towards, jacob2017quantization} and/or efficient architecture design~\cite{chollet2017xception, howard2017mobilenets}. 
\kui{Among them,} the quantization approaches represent weights and activations with low bitwidth fixed-point integers, \kui{and thus} the dot product can be computed by several \bohan{XNOR-popcount bitwise operations. The XNOR of two bits} \kui{requires only} a single logic gate instead of using hundreds units for floating point multiplication~\cite{ehliar2014area, govindu2004analysis}. Binarization~\cite{hubara2016binarized, rastegari2016xnor} is an extreme quantization approach where both the weights and activations are represented by a single bit, either +1 or -1. In this paper, we aim to design highly accurate \bohan{binary neural networks (BNNs)} from both the quantization and efficient architecture design perspectives. 

Existing quantization methods can be mainly divided into two categories. The first category methods seek to design more effective optimization algorithms to find better local minima for quantized weights.  \kui{These works} either introduce knowledge distillation~\cite{zhuang2018towards, polino2018model, mishra2018apprentice} or use loss-aware objectives~\cite{hou2018loss, hou2017loss}. 
The second category approaches focus on improving the quantization function~\cite{zhou2016dorefa, Cai_2017_CVPR, zhang2018lq}. \kui{To maintain good performance,} it is essential to learn suitable mappings between discrete values and their floating-point counterparts . However, \kui{designing quantization function} is highly non-trivial especially for BNNs, \kui{since the quantization functions are often non-differentiable and gradients can only be roughly approximated. }

The above two categories of methods belong to \emph{value approximation}, which seeks to quantize weights and/or activations by preserving most of the representational ability of the original network. 
However,  the value approximation approaches have a \kui{natural limitation that it is merely a local approximation}. Moreover, these methods \kui{often lacks of  adaptive ability} to general tasks. Given a pretrained model on a \kui{specific task, the quantization error will inevitably occur and  the final performance may be affected.}

\kui{In this paper, we seek to explore a third category called \emph{structure approximation}} . The main objective is to redesign a binary architecture that can \kui{directly} match the capability of \kui{a} floating-point model. In particular, we propose a Structured Binary Neural Network called Group-Net  to partition the full-precision model into groups and use a set of parallel binary bases to approximate its floating-point structure counterpart. 
In this way,   \kui{higher-level  structural information can be better preserved than  the \emph{value approximation} approaches}.   

\kui{What's more, relying on the proposed structured model, we are able to} design flexible binary structures according to different tasks and exploit \kui{task-specific information or structures} to compensate the quantization loss and facilitate training. 
For example, when transferring Group-Net from image classification to semantic segmentation, we are motivated by the structure of Atrous Spatial Pyramid Pooling (ASPP)~\cite{chen2018deeplab}. In  DeepLab v3~\cite{chen2017rethinking} and v3+~\cite{chen2018encoder}, ASPP is merely applied on the top of extracted features while each block in the backbone network can  employ one atrous rate only.
In contrast, we propose to directly apply different atrous rates on parallel binary bases in the backbone network, which is equivalent to absorbing ASPP into the feature extraction stage. In this way, we significantly boost \kui{the} performance on semantic segmentation, without increasing the computation complexity of the binary convolutions.

\kui{In general, it is nontrivial to extend previous \emph{value approximation} based quantization approaches to more challenging tasks such as semantic segmentation (or other general computer vision tasks)}. However, as will be shown, our Group-Net can be easily extended to other tasks. 
\kui{ Nevertheless, it is worth mentioning that value and structure approximation are complementary rather than contradictory.} In other words, \kui{both are important and should be  exploited to obtain highly accurate BNNs. }

Our methods are also motivated by those energy-efficient architecture design approaches~\cite{chollet2017xception, iandola2016squeezenet, howard2017mobilenets, zhang2017shufflenet} which seek to replace the traditional expensive convolution with computational efficient convolutional operations (\ie, depthwise separable convolution, $1 \times 1$ convolution). Nevertheless, we propose to design binary network architectures for dedicated hardware from the quantization view. \kui{We highlight that while most existing} quantization works focus on directly quantizing the full-precision architecture, at this point in time we do begin to explore alternative architectures \kui{ that shall be better suited for}  dealing with binary weights and activations.
In particular, apart from decomposing each group into several binary bases, we also propose to learn the connections between each group by introducing a fusion gate. \kui{ Moreover, Group-Net can be possibly further improved with  Neural Architecture Search methods~\cite{zoph2016neural, pham2018efficient, zoph2017learning} .}

Our contributions are summarized as follows:
	\begin{itemize}
	\itemsep -0.1cm
    \item{We propose to design accurate BNNs structures from the \emph{structure approximation} perspective. Specifically, we divide the networks into groups and approximate each group using a set of binary bases. We also propose to automatically learn the decomposition by introducing soft connections.}
    \item{
    \kui{The proposed Group-Net has strong flexibility and can be easily extended to other tasks. For instance, in this paper} we propose Binary Parallel Atrous Convolution (BPAC), which embeds rich multi-scale context into BNNs for accurate semantic segmentation. 
    Group-Net with BPAC significantly improves the performance while maintaining the complexity compared to employ Group-Net only.}
    \item{We evaluate our models on ImageNet and PASCAL VOC datasets based on ResNet. Extensive experiments show the proposed Group-Net achieves the state-of-the-art trade-off between accuracy and computational complexity.} 
	\end{itemize}

\section{Related Work}
\noindent\textbf{Network quantization}: The recent increasing demand for implementing fixed point deep neural networks on embedded devices motivates the study of network quantization. Fixed-point approaches that use low-bitwidth discrete values to approximate real ones have been extensively explored in the literature~\cite{Cai_2017_CVPR, zhou2016dorefa, hubara2016binarized, rastegari2016xnor, lin2017towards, zhuang2018towards}. BNNs~\cite{hubara2016binarized, rastegari2016xnor} propose to constrain both weights and activations to binary values (\ie, +1 and -1), where the multiply-accumulations can be replaced by purely $xnor(\cdot)$ and $popcount(\cdot)$ operations. To make a trade-off between accuracy and complexity, ~\cite{fromm2018heterogeneous, tang2017train, guo2017network, li2017performance} \bohan{propose to recursively perform residual quantization and yield a series of binary tensors with decreasing magnitude scales.} %
However, multiple binarizations are sequential process which cannot be paralleled. In~\cite{lin2017towards}, Lin \etal propose to use a linear combination of binary bases to approximate the floating point tensor during forward propagation. This inspires aspects of our approach, but unlike all of these local tensor approximations, we additionally directly design BNNs from a structure approximation perspective. 

\noindent\textbf{Efficient architecture design:} There has been a rising interest in designing efficient architecture in the recent literature. 
Efficient model designs like GoogLeNet~\cite{szegedy2015going} and SqueezeNet~\cite{iandola2016squeezenet} propose to replace 3$\times$3 convolutional kernels with 1$\times$1 size to reduce the complexity while increasing the depth and accuracy. Additionally, separable convolutions are also proved to be effective in Inception approaches~\cite{szegedy2016rethinking, szegedy2017inception}. This idea is further generalized as depthwise separable convolutions by Xception~\cite{chollet2017xception}, MobileNet~\cite{howard2017mobilenets, sandler2018mobilenetv2} and ShuffleNet~\cite{zhang2017shufflenet} to generate energy-efficient network structure.
To avoid handcrafted heuristics, neural architecture search~\cite{zoph2016neural, pham2018efficient, zoph2017learning, liu2017progressive, liu2017hierarchical} has been explored for automatic model design.

\section{Method}
\noindent \kui{Most previous literature has focused on value approximation by designing accurate binarization functions for weights and activations (\eg, multiple binarizations~\cite{lin2017towards, tang2017train, guo2017network, li2017performance, fromm2018heterogeneous}).} \kui{In this paper, we seek to binarize both  weights and activations of CNNs from a ``structure approximation" view.}   \kui{In the following, we first give the problem definition and some basics  about binarization in Sec.~\ref{sec:function}.} Then, in Sec.~\ref{sec:decomposition}, we explain our binary architecture design strategy. Finally, in Sec.~\ref{sec:segmentation}, we describe how to utilize task-specific attributes to generalize our approach to semantic segmentation.

\vspace{-2mm}
\subsection{Problem definition}\label{sec:function}
\noindent For a convolutional layer, we define the input ${\bf{x}} \in {\mathbb{R}^{{c_{in}} \times {w_{in}} \times {h_{in}}}}$, weight filter ${\bf{w}} \in {\mathbb{R}^{c \times w \times h}}$ and the output ${\bf{y}} \in {\mathbb{R}^{{c_{out}} \times {w_{out}} \times {h_{out}}}}$, respectively.

\noindent\textbf{Binarization of weights}: Following ~\cite{rastegari2016xnor}, we approximate the floating-point weight ${\bf{w}}$ by a binary weight filter ${{\bf{b}}^w}$ and a scaling factor $\alpha  \in {\mathbb{R}^ + }$ such that ${\bf{w}} \approx \alpha {{\bf{b}}^w}$, where ${{\bf{b}}^w}$ is the sign of ${\bf{w}}$ and $\alpha$ calculates the mean of absolute values of ${\bf{w}}$. In general, $sign(\cdot)$ is non-differentiable and so we adopt the straight-through estimator~\cite{bengio2013estimating} (STE) to approximate the gradient calculation. 
Formally, the forward and backward processes can be given as follows: 
\setlength{\abovedisplayskip}{1.5pt} 
\setlength{\belowdisplayskip}{1.5pt}
\begin{equation}
\begin{split}
&\mbox{Forward}:{{\bf{b}}^w} = \alpha  \cdot {\rm{sign}}({\bf{w}}),\\
&\mbox{Backward}:\frac{{\partial \ell }}{{\partial {\bf{w}}}} = \frac{{\partial \ell }}{{\partial {{\bf{b}}^w}}} \cdot \frac{{\partial {{\bf{b}}^w}}}{{\partial {\bf{w}}}} \approx \frac{{\partial \ell }}{{\partial {{\bf{b}}^w}}},\\
\end{split}
\end{equation}
where $\ell$ is the loss.

\noindent\textbf{Binarization of activations}: 
For activation binarization, we utilize the piecewise polynomial function to approximate the sign function as in~\cite{Liu_2018_ECCV}. The forward and backward can be written as:
\begin{equation}
\begin{array}{l}
\mbox{Forward}: {b^a} = \rm{sign}(x),\\
\mbox{Backward}:\frac{{\partial \ell }}{{\partial x}} = \frac{{\partial \ell }}{{\partial {b^a}}} \cdot \frac{{\partial {b^a}}}{{\partial x}},\\
\mbox{where} \quad \frac{{\partial {b^a}}}{{\partial x}} = \left\{ \begin{array}{l}
2 + 2x: - 1 \le x < 0\\
2 - 2x:0 \le x < 1\\
0:\rm{otherwise}
\end{array} \right..
\end{array}
\end{equation}

\subsection{Structured  Binary Network  Decomposition} \label{sec:decomposition}

\noindent \kui{In this paper, we seek to  design a new structural representation of a network for quantization. } \kui{First of all, note that a float number in computer is represented by a fixed-number of binary digits. Motivated by this, rather than directly doing the quantization via ``value decomposition", we propose to decompose a network  into binary structures while preserving its representability.} 

Specifically, given a floating-point residual network $\Phi$ with $N$ blocks,  we decompose $\Phi$ into $P$ binary fragments $[{{\cal F}_1}, ..., {{\cal F}_P}]$, 
where ${{\cal F}_i}(\cdot)$ can be any binary structure. Note that each ${{\cal F}_i}(\cdot)$ can be different. \kui{A natural question arises: can we find some simple  methods to decompose the network with binary structures so that the representability can be exactly preserved? To answer this question, we here explore two kinds of architectures}  for ${{\cal F}}(\cdot)$, namely layer-wise decomposition and group-wise decomposition in Sec.~\ref{sec:layerwise} and Sec.~\ref{sec:groupwise}, respectively. 
After that, we will present a novel strategy for automatic decomposition in Sec.~\ref{sec:learn}.
\begin{figure*}[!htb]
	\centering
	\resizebox{0.9\linewidth}{!}
	{
		\begin{tabular}{c}
			\includegraphics{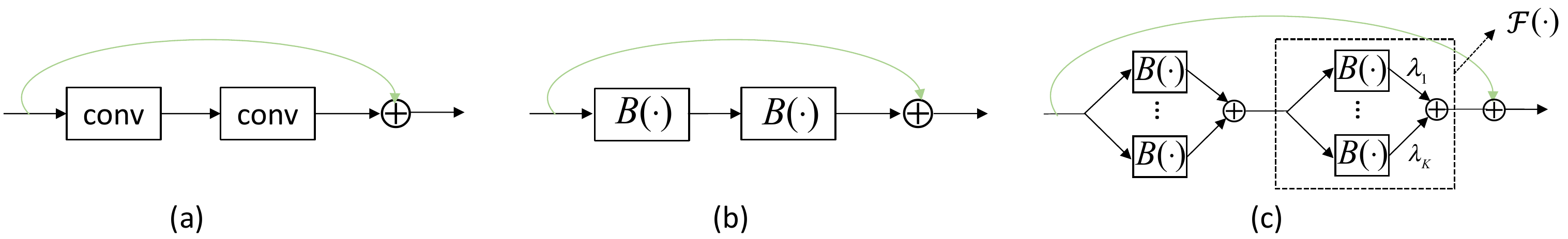}
		\end{tabular}
	}
	\caption{Overview of the baseline binarization method and the  proposed layer-wise binary decomposition. We take one residual block with two convolutional layers for illustration. For convenience, we omit batch normalization and nonlinearities. (a): The floating-point residual block. (b): \kui{Direct binarization of a full-precision block. (c): Layer-wise binary decomposition in Eq.~(\ref{eq:1}), where we use a set of binary convolutional layers $B(\cdot)$ to approximate a floating-point convolutional layer.}}
    \vspace{-1em}
	\label{fig:layerwise}
\end{figure*}

\subsubsection{Layer-wise binary decomposition} \label{sec:layerwise}
\vspace{-2mm}
The key challenge of binary decomposition is how to reconstruct or approximate the floating-point structure.  The simplest  way is to approximate in a layer-wise manner. Let $B(\cdot)$ be a binary convolutional layer and ${\bf{b}}_i^w$ be the binarized weights for the $i$-th layer. In Fig.~\ref{fig:layerwise} (c), we illustrate the layer-wise feature reconstruction for a single block.
Specifically, for each layer, we aim to fit the full-precision structure using a set of binarized homogeneous branches ${\cal F(\cdot)}$ given a floating-point input tensor $\bf{x}$:
\begin{equation}\label{eq:1}
{{\cal F}}({\bf{x}}) = \sum\limits_{i = 1}^K {{\lambda _i}{B_i}({\bf{x}})}  = \sum\limits_{i = 1}^K {{\lambda _i}} ({\bf{b}}_i^w \oplus sign({\bf{x}})),
\end{equation} 
where $\oplus$ is bitwise operations \textit{$xnor(\cdot)$} and \textit{$popcount(\cdot)$}, $K$ is the number of branches and ${\lambda_i}$ is the \kui{combination coefficient to be determined}. %
During the training, the structure is fixed and each binary convolutional kernel ${\bf{b}}_i^w$ as well as $\lambda_i$ are directly updated with end-to-end optimization.  The scale scalar can be absorbed into batch normalization when doing inference. 
\kui{Note that all ${{B_i}}$'s in Eq.~(\ref{eq:1}) have the same \bohan{topology} as the original floating-point counterpart.} Each binary branch gives a rough approximation and all the approximations are aggregated to achieve more accurate reconstruction to the original full precision convolutional layer. Note that when $K=1$, it corresponds to directly binarize the floating-point convolutional layer (Fig.~\ref{fig:layerwise} (b)). However,  with more branches (a larger $K$), we are expected to achieve more accurate approximation with more complex transformations. 

During the inference, the homogeneous $K$ bases can be parallelizable and thus the structure is hardware friendly. This will bring significant gain in  speed-up of the inference. 
Specifically, the bitwise XNOR operation and bit-counting can be performed in a parallel of 64 by the current generation of CPUs~\cite{rastegari2016xnor, Liu_2018_ECCV}. And we just need to calculate $K$ binary convolutions and $K$ full-precision additions. As a result, the speed-up ratio $\sigma$ for a convolutional layer can be calculated as:
\begin{equation}
\begin{aligned}
\sigma  &= \frac{{{c_{{\rm{in}}}}{c_{out}}wh{w_{in}}{h_{in}}}}{{\frac{1}{{64}}(K{c_{{\rm{in}}}}{c_{out}}wh{w_{in}}{h_{in}}) + K{c_{out}}{w_{out}}{h_{out}}}},\\
 &= \frac{{64}}{K} \cdot \frac{{{c_{{\rm{in}}}}wh{w_{in}}{h_{in}}}}{{{c_{{\rm{in}}}}wh{w_{in}}{h_{in}} + 64{w_{out}}{h_{out}}}}.
\end{aligned}
\end{equation}

\bohan{We take one layer in ResNet for example. If we set ${c_{in}} = 256$, $w \times h = 3 \times 3$, ${{\rm{w}}_{in}} = {h_{in}} = {w_{out}} = {h_{out}} = 28$, $K=5$, then it can reach 12.45$\times$ speedup. But in practice, each branch can be implemented in parallel. And the actual speedup ratio is also influenced by the process of memory read and thread communication. We further report real speedup ratio on CPU in Sec.~\ref{sec:hardware}.}
\vspace{-2mm}

\begin{figure}[!htb]
	\centering
	\resizebox{1.0\linewidth}{!}
	{
		\begin{tabular}{c}
			\includegraphics{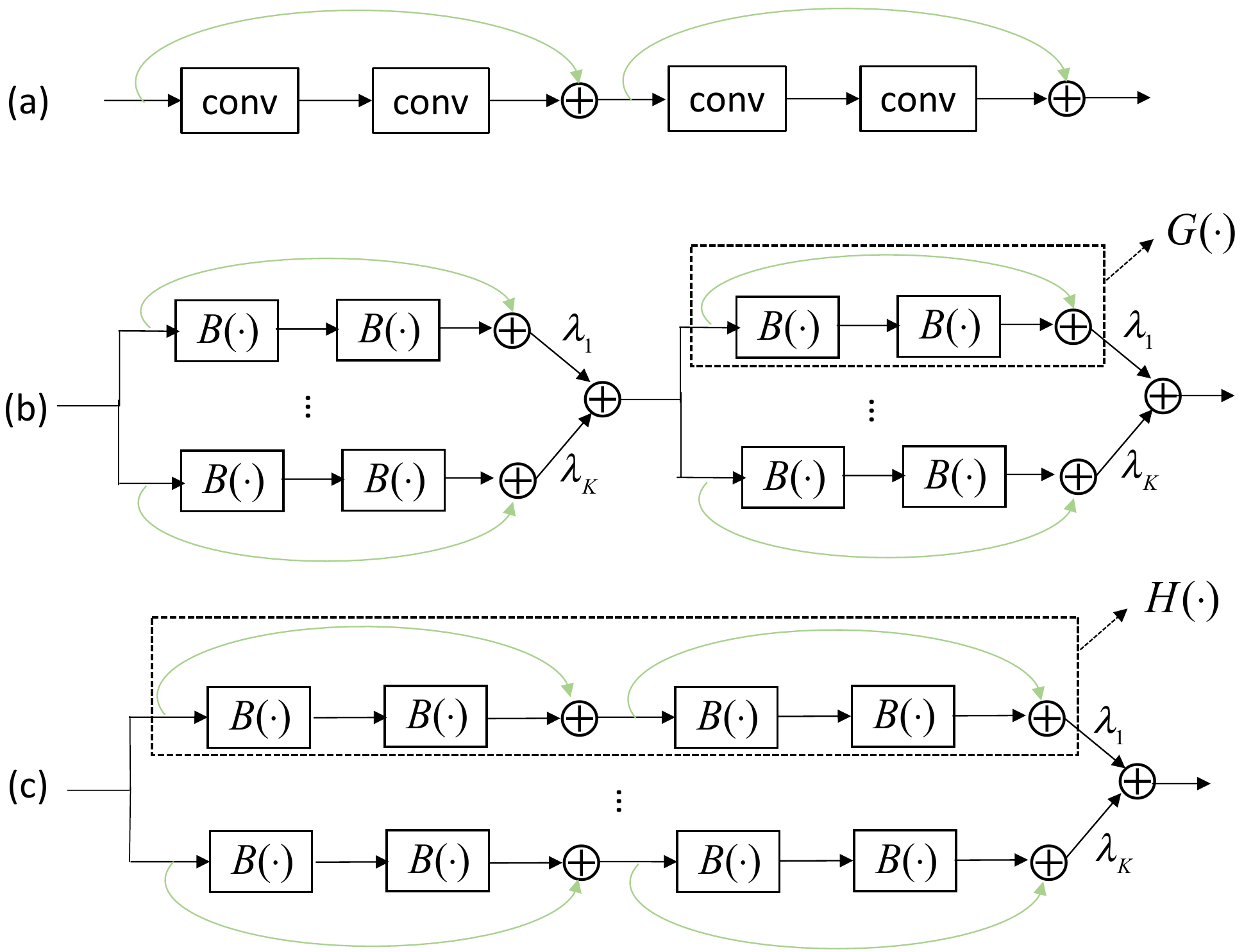}
		\end{tabular}
	}
	\caption{Illustration of the proposed group-wise binary decomposition strategy. We take two residual blocks for description. (a): The floating-point residual blocks. (b): Basic group-wise binary decomposition in Eq.~(\ref{eq:2}), where we approximate a whole block with a linear combination of binary blocks $G(\cdot)$. (c): We approximate a whole group with homogeneous binary bases $H(\cdot)$, where each group consists of several blocks. This corresponds to Eq.~(\ref{eq:3}).}
	\vspace{-1.0em}
	\label{fig:groupwise}
\end{figure}
\subsubsection{Group-wise binary decomposition}\label{sec:groupwise}
In \kui{the layer-wise approach, we approximate each layer with multiple branches of binary layers. Note each branch will introduce a certain amount of error and  the error may accumulate due to the aggregation of multi-branches. As a result, this strategy  may incur severe quantization errors and bring large deviation for gradients during backpropagation.} 
\bohan{To alleviate the above issue, we further propose a \kui{more flexible decomposition strategy called} group-wise binary decomposition, to preserve more structural information during approximation.}

\bohan{To explore the group-structure decomposition, we first consider \kui{a simple} case where each group consists of only one block. \kui{Then,} the  layer-wise approximation strategy can be easily extended to the group-wise case. As shown in Fig.~\ref{fig:groupwise} (b), similar to the layer-wise case, each floating-point group is decomposed into multiple binary groups. However,  each group ${G_i(\cdot)}$ is a binary block which consists of several binary convolutions and floating-point element-wise operations (\ie, ReLU, AddTensor). For example, we can set ${G_i(\cdot)}$ as the basic residual block~\cite{he2016deep} which is shown in Fig.~\ref{fig:groupwise} (a).}
Considering the \bohan{residual} architecture, we can decompose ${\cal F}({\bf{x}}) $  by extending Eq.~(\ref{eq:1}) as:
\setlength{\abovedisplayskip}{1pt} 
\setlength{\belowdisplayskip}{1pt}
\begin{equation} \label{eq:2}
{\cal F}({\bf{x}}) = \sum\limits_{i = 1}^K {{\lambda _i}{G _i}({\bf{x}})},
\end{equation}
where \kui{${\lambda_i}$ is the combination coefficient to be learned.} In Eq.~(\ref{eq:2}), we use a linear combination of homogeneous binary bases to approximate one group, where each base ${G _i}$ is a binarized block. 
In this way, we effectively keep the original residual structure in each base to preserve the network capacity. 
As shown in Sec.~\ref{sec:layervsgroup}, the group-wise decomposition strategy performs much better than the simple layer-wise approximation.

Furthermore, the group-wise approximation is flexible. We now analyze the case where each group may contain different number of blocks. Suppose we partition the network into $P$ groups and it follows a simple rule that each group must include one or multiple complete residual building blocks.
For the $p$-th group, we consider the blocks set $T \in \{ {T_{p - 1}} + 1,...,{T_p}\}$, where the index ${T_{p - 1}}=0$ if $p=1$. And we can extend Eq.~(\ref{eq:2}) into multiple blocks format:
\setlength{\abovedisplayskip}{1pt} 
\setlength{\belowdisplayskip}{1pt}
\begin{equation}\label{eq:3}
\begin{array}{l}
{\cal F}({{\bf{x}}_{{T_{p - 1}} + 1}}) = \sum\limits_{i = 1}^K {{\lambda _i}} {H_i}({\bf{x}}),\\
 = \sum\limits_{i = 1}^K {{\lambda _i}G_i^{{T_p}}(G_i^{{T_p} - 1}(...(G_i^{{T_{p - 1}} + 1}({{\bf{x}}_{{T_{p - 1}} + 1}}))...)} ),
\end{array}
\end{equation}
where $H(\cdot)$ is a cascade of consequent blocks which is shown in Fig.~\ref{fig:groupwise} (c).
Based on ${\cal{F(\cdot)}}$, we can efficiently construct a network by stacking these groups and each group may consist of one or multiple blocks. Different from Eq.~(\ref{eq:2}), we further expose a new dimension on each base, which is the number of blocks. This greatly increases the structure space and the flexibility of decomposition. %
We illustrate several possible connections in Sec. S1 in the supplementary file and further describe how to learn the decomposition in Sec.~\ref{sec:learn}.

\vspace{-3mm}
\subsubsection{Learning for dynamic decomposition} \label{sec:learn}
\noindent \kui{There is a big challenge involved in Eq.~(\ref{eq:3}). Note that the network has $N$ blocks and the possible number of connections is ${2^N}$. Clearly, it is not practical to enumerate all possible structures during the training. Here, we propose to solve this problem by  learning the structures for decomposition dynamically. We introduce in a fusion gate as the soft connection between blocks $G(\cdot)$. To this end, we first define the input of the $i$-th branch for the $n$-th block as}
\begin{equation} \label{eq:decompose}
\begin{aligned}
C_i^n &= \mathrm{sigmoid}({\theta ^n _i}),\\
{\bf{x}}_i^n&= {C ^n _i} \odot G_i^{n - 1}({\bf{x}}_i^{n - 1})  \\&+ (1 - {C^n_i}) \odot \sum\limits_{j = 1}^K {{\lambda _j}} G_j^{n - 1}({\bf{x}}_j^{n - 1}),
\end{aligned}
\end{equation}
where $\theta  \in {\mathbb{R}^K}$ is a learnable parameter vector and $C_i^n$ is a gate scalar. 

\begin{figure}[!htb]
	\centering
	\resizebox{1.0\linewidth}{!}
	{
		\begin{tabular}{c}
			\includegraphics{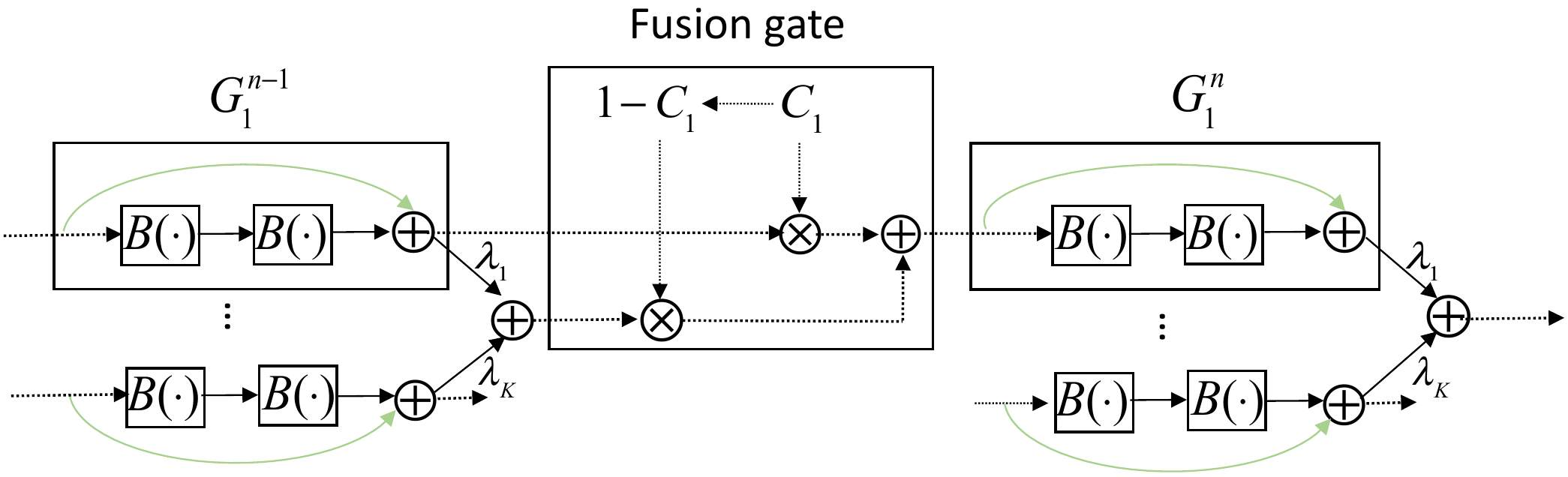}
		\end{tabular}
	}
	\caption{Illustration of the soft connection between two neighbouring blocks. For convenience, we only illustrate the fusion strategy for one branch.}
	\label{fig:decompose}
\end{figure}

Here, the branch input ${\bf{x}}_i^n$ is a weighted combination of two paths. The first path is the output of the corresponding $i$-th branch in the $(n-1)$-th block, which is a straight connection. The second path is the aggregation output of the $(n - 1)$-th block. The detailed structure is shown in Fig.~\ref{fig:decompose}. In this way, we make more information flow into the branch and increase the gradient paths for improving the convergence of BNNs.

\kui{\textbf{Remarks:} For the extreme case when {\small{$\sum\limits_{i = 1}^K {C_i^n} = 0$}}, Eq.~(\ref{eq:decompose}) will be reduced to Eq.~(\ref{eq:2}) which means we approximate the $(n-1)$-th and the $n$-th block independently.} When {\small{$\sum\limits_{i = 1}^K {{C_i^n}} = K$}},  Eq.~(\ref{eq:decompose})  is equivalent to Eq.~(\ref{eq:3})  and we set $H(\cdot)$ to be two consequent blocks and approximate the group as a whole.
Interestingly, when {\small{$\sum\limits_{n = 1}^N {\sum\limits_{i = 1}^K {C_i^n} } = NK$}}, it corresponds to set $H(\cdot)$ in Eq.~(\ref{eq:3}) to be a whole network and directly ensemble $K$ binary models. %
\begin{figure}[!htb]    
\centering
\resizebox{0.9\linewidth}{!}
{
	\begin{tabular}{c}
		\includegraphics{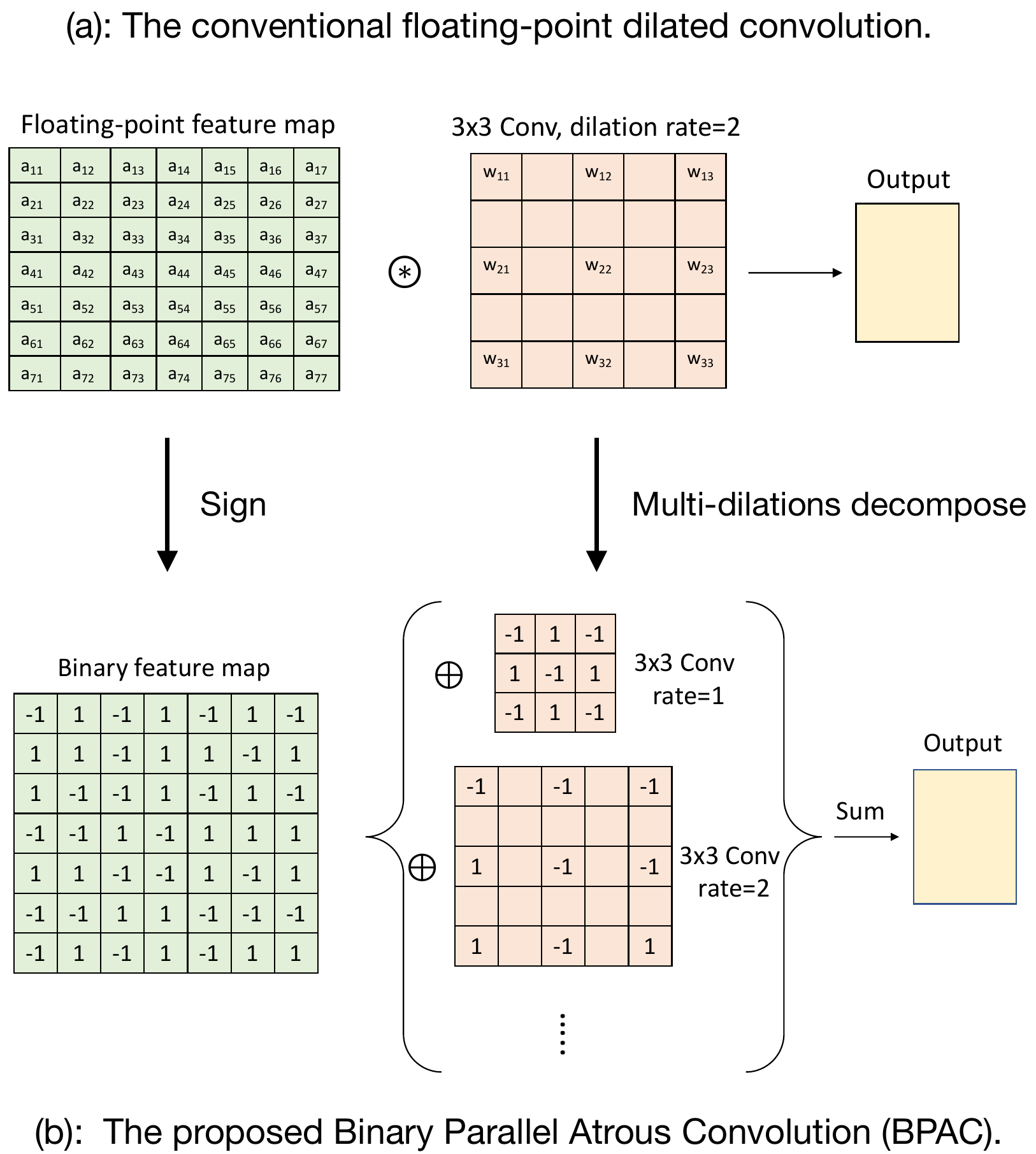}
	\end{tabular}
}
	
	\caption{The comparison between conventional dilated convolution and BPAC. For expression convenience, the group only has one convolutional layer. $\circledast$ is the convolution operation and $\oplus$ indicates the XNOR-popcount operations. (a): The original floating-point dilated convolution. (b): We decompose the floating-point atrous convolution into a combination of binary bases, where each base uses a different dilated rate. We sum the output features of each binary branch as the final representation.}
	\vspace{-1.0em}
	\label{fig:segmentation}
\end{figure}

\subsection{Extension to semantic segmentation} \label{sec:segmentation}
\noindent The key message conveyed in the proposed method is that although each binary branch has a limited modeling capability, aggregating them together leads to a powerful model. In this section, we show that this principle can be applied to tasks other than image classification. In particular, we consider semantic segmentation which can be deemed as a dense pixel-wise classification problem. In the state-of-the-art semantic segmentation network, the atrous convolutional layer~\cite{chen2017rethinking} is an important building block, which performs convolution with a certain dilation rate. To directly apply the proposed method to such a layer, one can construct multiple binary atrous convolutional branches with the same structure and aggregate results from them.  However, we choose not to do this but propose an alternative strategy: we use different dilation rates for each branch. In this way, the model can leverage multiscale information as \textit{a by-product of the network branch decomposition.} It should be noted that this scheme does not incur any additional model parameters and computational complexity compared with the naive binary branch decomposition. The idea is illustrated in Fig.~\ref{fig:segmentation} and we call this strategy Binary Parallel Atrous Convolution (\textbf{BPAC}). 

In this paper, we use the same ResNet backbone in~\cite{chen2017rethinking, chen2018encoder} with \emph{output\_stride=8}, where the last two blocks employ atrous convolution. In BPAC, we keep $rates = \{2,3,...,K,K+1\}$ and $rates = \{6,7,...,K+4,K+5\}$ for $K$ bases in the last two blocks, respectively.
Intriguingly, as will be shown in Sec.~\ref{exp:pascal}, our strategy brings so much benefit that \bohan{using five binary bases with BPAC}
achieves similar performance as the original full precision network despite the fact that it saves considerable computational cost. 

\section{Discussions} \label{sec:discussion}

\noindent\textbf{\kui{Complexity analysis}}: 
\begin{table*}[!htb]
	\centering
	\scalebox{0.8}
	{
		\begin{tabular}{c| c c c c c c c}
			\hline
			Model &Weights &Activations &Operations  &Memory saving &Computation Saving \\
			\hline
			Full-precision DNN &$F$  &$F$  &+, -, $\times$ &1  &1    \\\hline
			\cite{hubara2016binarized, rastegari2016xnor} &$B$ &$B$ &XNOR, popcount &$\sim 32 \times$ &$\sim 64 \times$   \\\hline
			\cite{courbariaux2015binaryconnect, hou2017loss} &$B$ &$F$  &+, - &$\sim 32$  &$\sim 2 \times $\\\hline
			\cite{zhu2016trained, zhou2017incremental} &$Q_K$  &$F$  &+, -, $\times$  &$\sim\frac{{32}}{K}$  &$< 2 \times$   \\\hline			
			\cite{zhang2018lq, zhuang2018towards, zhou2016dorefa, mishra2018apprentice} &$Q_K$ &$Q_K$ &+, -, $\times$  & $\sim\frac{{32}}{K}$ &$< \frac{{64}}{{{K^2}}} \times$  \\\hline
			\cite{lin2017towards, fromm2018heterogeneous, tang2017train, guo2017network, li2017performance} &$K \times B$  &$K \times B $ & +, -, XNOR, popcount &$\sim\frac{{32}}{K}$ & $ < \frac{{64}}{{{K^2}}} \times$  \\\hline
		    Group-Net &\multicolumn{2}{|c}{$K \times (B, B)$} &+, -, XNOR, popcount  &$\sim\frac{{32}}{K}$ &$< \frac{{64}}{K} \times$ \\\hline 
			
	\end{tabular}}
	\caption{Computational complexity and storage comparison  of different quantization approaches. $F$: full-precision, $B$: binary, $Q_K$: $K$-bit quantization.}
	\vspace{-1em}
	\label{tab:complexity_compare}
\end{table*}
A comprehensive comparison of various quantization approaches over complexity and storage is shown in Table~\ref{tab:complexity_compare}. For example, in the previous state-of-the-art binary model ABC-Net~\cite{lin2017towards}, each convolutional layer is approximated using $K$ weight bases and $K$ activation bases, which needs to calculate  $K^2$ times binary convolution. In contrast, we just need to approximate several groups with $K$ structural bases. As reported in Sec.~\ref{sec:imagenet} , we save approximate $K$ times computational complexity while still achieving comparable Top-1 accuracy.
Since we use $K$ structural bases, the number of parameters increases by $K$ times in comparison to the full-precision counterpart. But we still save memory bandwidth by $32/K$ times since all the weights are binary in our paper.
For our approach, there exists element-wise operations between each group, so the computational complexity saving is slightly less than $\frac{{64}}{K} \times$.

\noindent\textbf{\kui{Differences of Group-net from fixed-point methods}}: The proposed \kui{Group-net} with $K$ bases is different from the $K$-bit fixed-point approaches~\cite{zhang2018lq, zhuang2018towards, zhou2016dorefa, mishra2018apprentice}. 

\kui{We first show how the inner product between fixed-point weights and activations can be computed by bitwise operations.
Let a weight vector ${\bf{w}} \in {\mathbb{R}^M}$ be encoded by a vector ${\bf{b}}_i^w \in {\{  - 1,1\} ^M}$, $i = 1,...,K$. Assume we also quantize activations to $K$-bit. Similarly, the activations $\bf{x}$ can be encoded by ${\bf{b}}_j^a \in {\{  - 1,1\} ^M}$, $j = 1,...,K$. Then, the convolution can be written as}
\begin{equation} \label{eq:fixed-point}
{Q_K}({{\bf{w}}^T}){Q_K}({\bf{x}}) = \sum\limits_{i = 0}^{K-1} {\sum\limits_{j = 0}^{K-1} {{2^{i + j}}} } ({\bf{b}}_i^w \xor {\bf{b}}_j^a),
\end{equation}
where $Q_K(\cdot)$ is \kui{any} quantization function\footnote{\noindent \kui{For simplicity, we only consider uniform quantization in this paper.}}. 

\kui{During the inference, it needs to first get} the encoding ${\bf{b}}_j^a$ for each bit via linear regression. \kui{Then, it calculates} and sums over $K^2$ times \textit{$xnor(\cdot)$} and \textit{$popcount(\cdot)$}. \kui{The complexity is  about $O(K^2)$}. \kui{Note that the output range} for a single output shall be $[ - {({2^K} - 1)^2}M,  {({2^K} - 1)^2}M]$. 

\kui{In contract, we directly obtain ${\bf{b}}_j^a$ via $sign(\bf{x})$.}
Moreover, \kui{since we just need to calculate $K$ times \textit{$xnor(\cdot)$} and  \textit{$popcount(\cdot)$} (see Eq.~(\ref{eq:1})), and then sum over the outputs, the computational complexity is $O(K)$.} For binary convolution, its output range is \{-1, 1\}. So the value range for each element after summation is $[ - KM, KM]$, in which the number of distinct values  is much less than that in fixed-point methods. 

\kui{In summary, \kui{compared with $K$-bit fixed-point methods,} Group-Net with $K$ bases just needs $\sqrt K$ computational complexity and saves ${({2^K} - 1)^2}/K$ accumulator bandwidth.} \kui{Even $\sqrt K $-bit fixed-point quantization requires more memory bandwidth to feed signal in SRAM or in registers.}

\noindent\textbf{\kui{Differences of Group-net from multiple binarizations methods}:}
In ABC-Net~\cite{lin2017towards}, a linear combination of binary weight/activations bases are obtained from the full-precision weights/activations without being directly learned. In contrast, we directly design the binary network structure, where binary weights are end-to-end optimized. ~\cite{fromm2018heterogeneous, tang2017train, guo2017network, li2017performance} propose to recursively approximate the residual error and obtain a series of binary maps corresponding to different quantization scales.
However, it is a sequential process which cannot be paralleled. And all multiple binarizations methods belong to local tensor approximation. 
In contrast to value approximation, we propose a structure
approximation approach to mimic the full-precision network. Moreover, tensor-based methods are tightly designed to local value approximation and are hardly generalized to other tasks accordingly. In addition, our structure decomposition strategy achieves much better performance than tensor-level approximation as shown in Sec.~\ref{sec:layervsgroup}.
More discussions are provided in Sec. S2 in the supplementary file.

\section{Experiment}
\noindent{We define several methods for comparison as follows:}
\noindent\textbf{LBD:} It implements the layer-wise binary decomposition strategy described in Sec.~\ref{sec:layerwise}.
\noindent\textbf{Group-Net:} It implements the full model with learnt soft connections described in Sec.~\ref{sec:learn}. 
\noindent\textbf{Group-Net**:} Following Bi-Real Net~\cite{Liu_2018_ECCV}, we apply shortcut bypassing every binary convolution to improve the convergence. And this strategy can only be applied on basic blocks in ResNet-18 and ResNet-34. 

\subsection{Implementation details}
\noindent As in \cite{rastegari2016xnor, Cai_2017_CVPR, zhou2016dorefa, zhuang2018towards}, we quantize the weights and activations of all convolutional layers except that the first and the last layer have full-precision. In all ImageNet experiments, training images are resized to $256 \times 256$, and a $224 \times 224$ crop is randomly sampled from an image or its horizontal flip, with the per-pixel mean subtracted. We do not use any further data augmentation in our implementation. We use a simple single-crop testing for standard evaluation. No bias term is utilized.
We first pretrain the full-precision model as initialization and fine-tune the binary counterpart.\footnote{For pretraining \emph{Group-Net}, we use $ReLU(\cdot)$ as nonlinearity. For pretraining \emph{Group-Net**}, we use $Tanh(\cdot)$ as nonlinearity.} We use Adam~\cite{kingma2014adam} for optimization. For training all binary networks, the mini-batch size and weight decay are set to 256 and 0, respectively. The learning rate starts at 5e-4 and is decayed twice by multiplying 0.1 at the 30th and 40th epoch. We train 50 epochs in total. Following~\cite{Cai_2017_CVPR, zhuang2018towards}, no dropout is used due to binarization itself can be treated as a regularization.
We apply layer-reordering to the networks as: Sign $\to$ Conv $\to$ ReLU  $\to$ BN. Inserting $ReLU(\cdot)$ after convolution is important for convergence.
Our simulation implementation is based on Pytorch~\cite{paszke2017automatic}. 

\vspace{-2mm}
\subsection{Evaluation on ImageNet} \label{sec:imagenet}
\noindent The proposed method is evaluated on ImageNet (ILSVRC2012) \cite{russakovsky2015imagenet} dataset. ImageNet is a large-scale dataset which has $\sim$1.2M training images from 1K categories and 50K validation images. Several representative networks are tested: ResNet-18~\cite{he2016deep}, ResNet-34 and ResNet-50.
As discussed in Sec.~\ref{sec:discussion}, binary approaches and fixed-point approaches differ a lot in computational complexity as well as storage consumption. So we compare the proposed approach with binary neural networks in Table~\ref{tab:binary_compare} and fixed-point approaches in Table~\ref{tab:fixed-point}, respectively.

\vspace{-4mm}
\subsubsection{Comparison with binary neural networks} \label{sec:compare_binary}

Since we employ binary weights and binary activations, we directly compare to the previous state-of-the-art binary approaches, including BNN~\cite{hubara2016binarized}, XNOR-Net~\cite{rastegari2016xnor}, Bi-Real Net~\cite{Liu_2018_ECCV} and ABC-Net~\cite{lin2017towards}. 
We report the results in Table~\ref{tab:binary_compare} and summarize the following points. 1): The most comparable baseline for Group-Net is ABC-Net. As discussed in Sec.~\ref{sec:discussion}, we save considerable computational complexity while still achieving better performance compared to ABC-Net. In comparison to directly binarizing networks, Group-Net achieves much better performance but needs $K$ times more storage and complexity. However, the $K$ homogeneous bases can be easily parallelized on the real chip. In summary, our approach achieves the best trade-off between computational complexity and prediction accuracy.
2): By comparing \emph{Group-Net** (5 bases)} and \emph{Group-Net (8 bases)}, we can observe comparable performance. \emph{It justifies adding more shortcuts can facilitate BNNs training} which is consistent with~\cite{Liu_2018_ECCV}. 
3): For Bottleneck structure in ResNet-50, we find larger quantization error than the counterparts using basic blocks with $3 \times 3$ convolutions in ResNet-18 and ResNet-34. The similar observation is also claimed by~\cite{bethge2018learning}. We assume that this is mainly attributable to the $1 \times 1$ convolutions in Bottleneck. The reason is $1 \times 1$ filters are limited to two states only (either 1 or -1) and they have very limited learning power. What's more, 
the bottleneck structure reduces the number of filters significantly, which means the gradient paths are greatly reduced. In other words, it blocks the gradient flow through BNNs. 
 \emph{Even though the bottleneck structure can benefit full-precision training, it is really needed to be redesigned in BNNs. To increase gradient paths, the $1 \times 1$ convolutions should be removed.}

\begin{table*}[ht]
	\centering
	\scalebox{0.7}
	{
		\begin{tabular}{c| c | c c c c c c  c c}
			\hline
			\multicolumn{2}{c|}{Model} &Full &BNN &XNOR &Bi-Real Net &ABC-Net (25 bases) &Group-Net (5 bases) &Group-Net** (5 bases) & Group-Net (8 bases) \\\hline
			\hline
			\multirow{2}{*}{ResNet-18}& Top-1 \% &69.7  &42.2  &51.2  &56.4 &65.0 &64.8 &67.0 &{\bf{67.5}}  \\ 
			&Top-5 \% &89.4  &67.1  &73.2  &79.5  &85.9 &85.7  &87.5 &{\bf88.0}  \\\hline
			\multirow{2}{*}{ResNet-34}& Top-1 \% &73.2   &-  &-  &62.2 &68.4  &68.5 &70.5  &{\bf{71.8}} \\ 
            &Top-5 \% &91.4  &-   &-  &83.9  &88.2  &88.0  &89.3  &\bf{90.4}   \\\hline
			\multirow{2}{*}{ResNet-50}& Top-1 \% &76.0  &-  &-  &-  &70.1 &69.5 &-  &{\bf{72.8}} \\ 
			&Top-5 \% &92.9  &-  &-  &-  &89.7  &89.2 &-  &{\bf{90.5}} \\\hline    	 
			
	\end{tabular}}
	\caption{Comparison with the state-of-the-art binary models using ResNet-18, ResNet-34 and ResNet-50 on ImageNet. All the comparing results are directly cited from the original papers. The metrics are Top-1 and Top-5 accuracy.}
    \vspace{-1.0em}
	\label{tab:binary_compare}
\end{table*}

\vspace{-5pt}
\subsubsection{Comparison with fix-point approaches} \label{sec:fixed-point}

Since we use $K$ binary group bases, we compare our approach with at least $\sqrt K$-bit fix-point approaches. In Table~\ref{tab:fixed-point}, we compare our approach with the state-of-the-art fixed-point approaches DoReFa-Net~\cite{zhou2016dorefa}, SYQ~\cite{faraonesyq} and LQ-Nets~\cite{zhang2018lq}.  As described in Sec.~\ref{sec:discussion}, $K$ binarizations are more superior than the $\sqrt K$-bit width quantization with respect to the resource consumption. Here, we set $K$=4. DOREFA-Net and LQ-Nets use 2-bit weights and 2-bit activations. SYQ employs binary weights and 8-bit activations. All the comparison results are directly cited from the corresponding papers. %
LQ-Nets is the current best-performing fixed-point approach and its activations have a long-tail distribution. We can observe that Group-Net requires less memory bandwidth while still achieving comparable accuracy with LQ-Nets.

\begin{table}[ht]
	\centering
	\scalebox{0.75}
{
	\begin{tabular}{c c c c c}
		\hline
		Model &W &A &Top-1 (\%)&Top-5 (\%) \\\hline
		Full-precision &32 &32  &69.7 &89.4   \\
		Group-Net** (4 bases)&1 &1 &\bf{66.3} &\bf{86.6} \\
		Group-Net (4 bases) &1 &1 &64.2  &85.6 \\
		LQ-Net~\cite{zhang2018lq} & 2 & 2 &64.9  &85.9 \\
		DOREFA-Net~\cite{zhou2016dorefa} &2 &2 &62.6 &84.4 \\
		SYQ~\cite{faraonesyq} &1 &8 &62.9 & 84.6 \\\hline
\end{tabular}}
	\caption{Comparison with the state-of-the-art fixed-point models with ResNet-18 on ImageNet. The metrics are Top-1 and Top-5 accuracy.}
	\vspace{-1.0em}
	\label{tab:fixed-point}
\end{table}

\begin{table}[h]
	\centering
	\scalebox{0.75}
	{
		\begin{tabular}{c c |c c}
			\hline
			\multicolumn{2}{c|}{Model} &mIOU & $\Delta$\\
			\hline
			\multirow{5}{*}{ResNet-18, FCN-32s}& Full-precision &64.9 &- \\
			&LQ-Net (3-bit) &62.5  &2.4 \\
			&Group-Net &60.5 &4.4 \\
			&Group-Net + BPAC  &63.8 &1.1 \\
			&Group-Net** + BPAC & \bf{65.1} &\bf{-0.2} \\\hline
			
			\multirow{5}{*}{ResNet-18, FCN-16s}& Full-precision & 67.3 &-\\ 
			&LQ-Net (3-bit) &65.1  &2.2 \\
			&Group-Net  &62.7 &4.6 \\
			&Group-Net + BPAC  &66.3 &1.0  \\
			&Group-Net** + BPAC &\bf{67.7} &\bf{-0.4} \\\hline		
			\multirow{5}{*}{ResNet-34, FCN-32s}& Full-precision &72.7 &- \\ 
		   &LQ-Net (3-bit) &70.4 &2.3 \\
           &Group-Net  &68.2  &4.5 \\
           &Group-Net + BPAC  &71.2 &1.5    \\
           &Group-Net** + BPAC &\bf{72.8} &\bf{-0.1} \\\hline					 
		    \multirow{4}{*}{ResNet-50, FCN-32s}& Full-precision &73.1  &-\\ 
		    &LQ-Net (3-bit) &{\bf{70.7}}  &{\bf{2.4}} \\
			&Group-Net &67.2  &5.9  \\
			&Group-Net + BPAC &70.4 &2.7\\\hline
			
	\end{tabular}}
	\caption{Performance on PASCAL VOC 2012 validation set.}
	\vspace{-1.0em}
	\label{tab:segmentation}
\end{table}

\vspace{-4mm}
\subsubsection{Hardware implementation on ImageNet} \label{sec:hardware}
We currently implement ResNet-34 using \emph{Group-Net**} and test the inference time on XEON E5-2630 v3 CPU with 8 cores. We use the off-the-shelf BMXNet~\cite{yang2017bmxnet} and OpenMP for acceleration.
The speedup ratio for convolution using \emph{xnor\_64\_omp} can reach more than 100$\times$. Moreover, $BN + sign(\cdot)$ is accelerated by XNOR operations following the implementation in FINN~\cite{umuroglu2017finn}.
We cannot accelerate floating-point element-wise operations including tensor additions and $ReLU(\cdot)$. We finally achieve {$\sim$\bf{7.5x}} speedup in comparison to the original ResNet-34 model based on MXNet~\cite{chen2015mxnet}, where thread communication occupies a lot of time in CPU. We also achieve actual {$\sim$\bf{5.8x}} memory saving. However, we expect to achieve better acceleration on parallelization friendly FPGA platforms but currently we do not have enough resources.

\subsection{Ablation study}
\noindent Due to the limited space, we provide more experiments in Sec. S1 in the supplementary material. 
\vspace{-3mm}
\subsubsection{Layer-wise vs. group-wise binary decomposition} \label{sec:layervsgroup}
We explore the difference between layer-wise and group-wise design strategies in Table~\ref{tab:groupspace}. By comparing the results, we find \emph{Group-Net} outperforms \emph{LBD} by 7.2\% on the Top-1 accuracy. Note that \emph{LBD} approach can be treated as a kind of \emph{tensor approximation} which has similarities with multiple binarizations methods in~\cite{fromm2018heterogeneous, guo2017network,  li2017performance, lin2017towards, tang2017train} and the differences are described in Sec. 4. It strongly shows the necessity for employing the group-wise decomposition strategy to get promising results. We speculate that this significant gain is partly due to the preserved block structure in binary bases. 
It also proves that apart from designing accurate binarization function, it is also essential to design appropriate structure for BNNs. 
\begin{table}[h]
	\centering
	\scalebox{0.85}
	{
		\begin{tabular}{c c c c}
			\hline
			Model &Bases &Top-1 \%&Top-5 \%\\\hline
		    Full-precision &1   &69.7 &89.4  \\
		    Group-Net &5  &\bf{64.8}  &\bf{85.7}\\
			LBD &5 &57.6 &79.7 \\
			\hline
	\end{tabular}}
	\caption{Comparison with \emph{Group-Net} and \emph{LBD} using ResNet-18 on ImageNet. The metrics are Top-1 and Top-5 accuracy.}
	\label{tab:groupspace}
\end{table}

\vspace{-4mm}
\subsection{Evaluation on PASCAL VOC} \label{exp:pascal}
\noindent We evaluate the proposed methods on the PASCAL VOC 2012 semantic segmentation benchmark~\cite{everingham2010pascal} which contains 20 foreground object classes and one background class. 
The original dataset contains 1,464 (\emph{train}), 1,449 (\emph{val}) and 1,456 (\emph{test}) images. 
The dataset is augmented by the extra annotations from~\cite{hariharan2011semantic}, resulting in 10,582 training images. The performance is measured in terms of averaged pixel intersection-over-union (mIOU) over 21 classes. Our experiments are based on \bohan{the original} FCN~\cite{long2015fully}. For both FCN-32s and FCN-16s, we adjust the dilation rates of the last 2 blocks in ResNet with atrous convolution to make the output stride equal to 8. We first pretrain the binary backbone network on ImageNet dataset and fine-tune it on PASCAL VOC. During fine-tuning, we use Adam with initial learning rate=1e-4, weight decay=1e-5 and batch size=16. We set the number of bases $K=5$ in experiments. We train 40 epochs in total and decay the learning rate by a factor of 10 at 20 and 30 epochs. We do not add any auxiliary loss and ASPP. 
We empirically observe full-precision FCN under dilation rates (4, 8) in last two blocks achieves the best performance. The main results are in Table~\ref{tab:segmentation}. 

From the results, we can observe that when all bases using the same dilation rates, there is a large performance gap with the full-precision counterpart. This performance drop \textbf{is consistent with} the classification results on ImageNet dataset in Table~\ref{tab:binary_compare}. It proves that the quality of extracted features have a great impact on the segmentation performance. What's more, by utilizing task-specific BPAC, we find significant performance increase with no computational complexity added, which strongly justifies the flexibility of Group-Net. Moreover, we also quantize the backbone network using fixed-point LQ-Nets with 3-bit weights and 3-bit activations. Compared with LQ-Nets, we can achieve comparable performance while saving considerable complexity.
In addition, we can observe \emph{Group-Net + BPAC} based on ResNet-34 even outperform the counterpart on ResNet-50. This shows the widely used bottleneck structure is not suited to BNNs as explained in Sec.~\ref{sec:compare_binary}.
We provide more analysis in Sec. S3 in the supplementary file.

\section{Conclusion}

\noindent In this paper, we have begun to explore highly efficient and accurate CNN architectures with binary weights and activations. Specifically, we have proposed to directly decompose the full-precision network into multiple groups and each group is approximated using a set of binary bases which can be optimized in an end-to-end manner. We also propose to learn the decomposition automatically.
Experimental results have proved the effectiveness of the proposed approach on the ImageNet classification task.
Moreover, we have generalized Group-Net from image classification task to semantic segmentation and achieved promising performance on PASCAL VOC. 
We have implemented the homogeneous multi-branch structure on CPU and achieved promising acceleration on test-time inference.

\def\kuired{\textcolor{red}}
\def\diag{\mbox{diag}}
\def\rank{\mbox{rank}}
\def\grad{\mbox{\text{grad}}}
\def\dist{\mbox{dist}}
\def\sgn{\mbox{sgn}}
\def\tr{\mbox{tr}}
\def\etal{{\em et al.\/}\, }
\def\card{{\mbox{Card}}}
\def\st{\mbox{s.t. }}

\renewcommand{\thesection}{S\arabic{section}}
\renewcommand{\thetable}{S\arabic{table}}
\renewcommand{\thefigure}{S\arabic{figure}}
\def\bx{ {\bf{\widetilde{x}}} }
\def\bs{ {\bf{\widetilde{s}}} }

\section*{Appendix}

\section{More ablation study on ImageNet classification} \label{sec:supply_s1}
\begin{figure*}[!htb]
	\centering
	\resizebox{1.0\linewidth}{!}
	{
		\begin{tabular}{c}
			\includegraphics{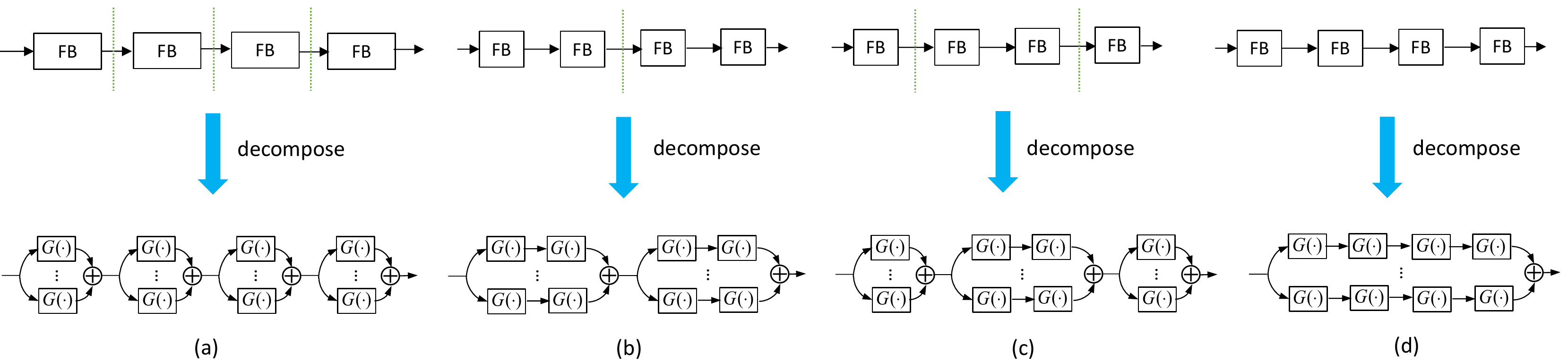}
		\end{tabular}
	}
	\caption{Illustration of several possible group-wise architectures. We assume the original full-precision network comprises four blocks. ``FB'' represents the floating-point block. $G(\cdot)$ is defined in Sec. 3.2.2 in the main paper, which represents a binary block.
	We omit the skip connections for convenience. (a): Each group comprises one block and we approximate each floating-point block with a set of binarized blocks. (b): Decompose the network into groups, where each group contains two blocks. Then we approximate each floating-point group using a set of binarized groups.
	(c): Each group contains different number of blocks. (d): An extreme case. We directly decompose the whole floating-point network into an ensemble of several binary networks.}
	\vspace{-1.0em}
	\label{fig:framework}
\end{figure*}

In this section, we continue the Sec. 5.3 in the main paper to provide more comparative experiments.
We define more methods for comparison as follows:
\noindent\textbf{GBD v1:} We implement with the group-wise binary decomposition strategy, where each base consists of one block. It corresponds to the approach described in Eq. (5) and is illustrated in Fig.~\ref{fig:framework} (a).
\noindent\textbf{GBD v2:} Similar to \emph{GBD v1}, the only difference is that each group base has two blocks. It is illustrated in Fig.~\ref{fig:framework} (b) and is explained in Eq. (6).
\noindent\textbf{GBD v3:} It is an extreme case where each base is a whole network, which can be treated as an ensemble of a set of binary networks. This case is shown in Fig.~\ref{fig:framework} (d).

\begin{table*}[h]
	\centering
	\scalebox{0.85}
	{
		\begin{tabular}{c c c c c c}
			\hline
			Model &Bases &Top-1 \%&Top-5 \%&Top-1 gap \% &Top-5 gap \% \\\hline
			ResNet-18 Full-precision &1   &69.7 &89.4 &- &-  \\
			Group-Net &5  &64.8  &85.7  &4.9  &3.7\\
			GBD v1 &5 &63.0 &84.8 & 6.7 &4.6 \\
		    GBD v2 &5 &62.2 &84.1 & 7.5 &5.3 \\
			GBD v3 &5 &59.2 &82.3 &10.5 &7.1 \\
			\hline
	\end{tabular}}
	\caption{Comparisons between several group-wise decomposition strategies. Top-1 and Top-5 accuracy gap to the corresponding full-precision networks are also reported.}
	\label{tab:groupspace}
\end{table*}
\subsection{Group space exploration}
We are interested in exploring the influence of different group-wise decomposition strategies. We present the results in Table~\ref{tab:groupspace}. We observe that by learning the soft connections between each block results in the best performance on ResNet-18. And methods based on hard connections perform relatively worse.  From the results, we can conclude that designing compact binary structure is essential for highly accurate classification. What's more, we expect to further boost the performance by integrating with the NAS approaches as discussed in Sec.~\ref{sec:supply_s2}.

\subsection{Effect of the number of bases} \label{sec:bases}
\begin{table*}[!htb]
	\centering
	\scalebox{0.95}
	{
		\begin{tabular}{c c c c c c c}
			\hline
			Model &Bases &Top-1 \%&Top-5 \%&Top-1 gap \% &Top-5 gap \% \\\hline
			Full-precision &1  &69.7 &89.4 &- &-  \\
			Group-Net &1  &55.6  &78.6  &14.1  &10.8 \\
			Group-Net &3  &62.5 &84.2 &7.2 &5.2  \\
			Group-Net &5  & 64.8  &85.7  &4.9 &3.7 \\\hline
	\end{tabular}}
	\caption{Validation accuracy of Group-Net on ImageNet with different number of bases. All cases are based on the ResNet-18 network with binary weights and activations.}
	\label{tab:paper_bases}
\end{table*}
We further explore the influence of number of bases $K$ to the final performance in Table~\ref{tab:paper_bases}. When the number is set to 1, it corresponds to directly binarize the original full-precision network and we observe apparent accuracy drop compared to its full-precision counterpart. With more bases employed, we can find the performance steadily increases. The reason can be attributed to the better fitting of the floating-point structure, which is a trade-off between accuracy and complexity. It can be expected that with enough bases, the network should has the capacity to approximate the full-precision network precisely. With the multi-branch group-wise design, we can achieve high accuracy while still significantly reducing the inference time and power consumption. Interestingly, each base can be implemented using small resource and the parallel structure is quite friendly to FPGA/ASIC.

\section{More discussions} \label{sec:supply_s2}

\noindent\textbf{Relation to ResNeXt~\cite{xie2017aggregated}}:
The homogeneous multi-branch architecture design shares some spirit of ResNeXt and enjoys the advantage of introducing a ``cardinality'' dimension. However, our objectives are totally different. ResNeXt aims to increase the capacity while maintaining the complexity. To achieve this, it first divides the input channels into groups and perform efficient group convolutions implementation. Then all the group outputs are aggregated to approximate the original feature map. In contrast, we first divide the network into groups and directly replicate the floating-point structure for each branch while both weights and activations are binarized. In this way, we can reconstruct the full-precision structure via aggregating a set of low-precision transformations for complexity reduction in the energy-efficient hardware. Furthermore, our structured transformations are not restricted to only one block as in ResNeXt.

\noindent\textbf{Group-Net has strong flexibility}:
The group-wise approximation approach can be efficiently integrated with Neural Architecture Search (NAS) frameworks~\cite{zoph2016neural,  pham2018efficient, liu2018darts, zoph2017learning, liu2017progressive} to explore the optimal architecture. Based on Group-Net, we can further add number of bases, filter numbers, connections among bases into the search space. The proposed approach can also be combined with knowledge distillation strategy as in~\cite{zhuang2018towards, mishra2018apprentice}. The basic idea is to train a target low-precision network alongside another pretrained full-precision guidance network. An additional regularizer is added to minimize the difference between student's and teacher's intermediate feature representations for higher accuracy. In this way, we expect to further decrease the number of bases while maintaining the performance.

\section{More ablation study on semantic segmentation} \label{sec:supply_s3}

\begin{figure}[!htb]
	\centering
	\resizebox{1.0\linewidth}{!}
	{
		\begin{tabular}{c}
			\includegraphics{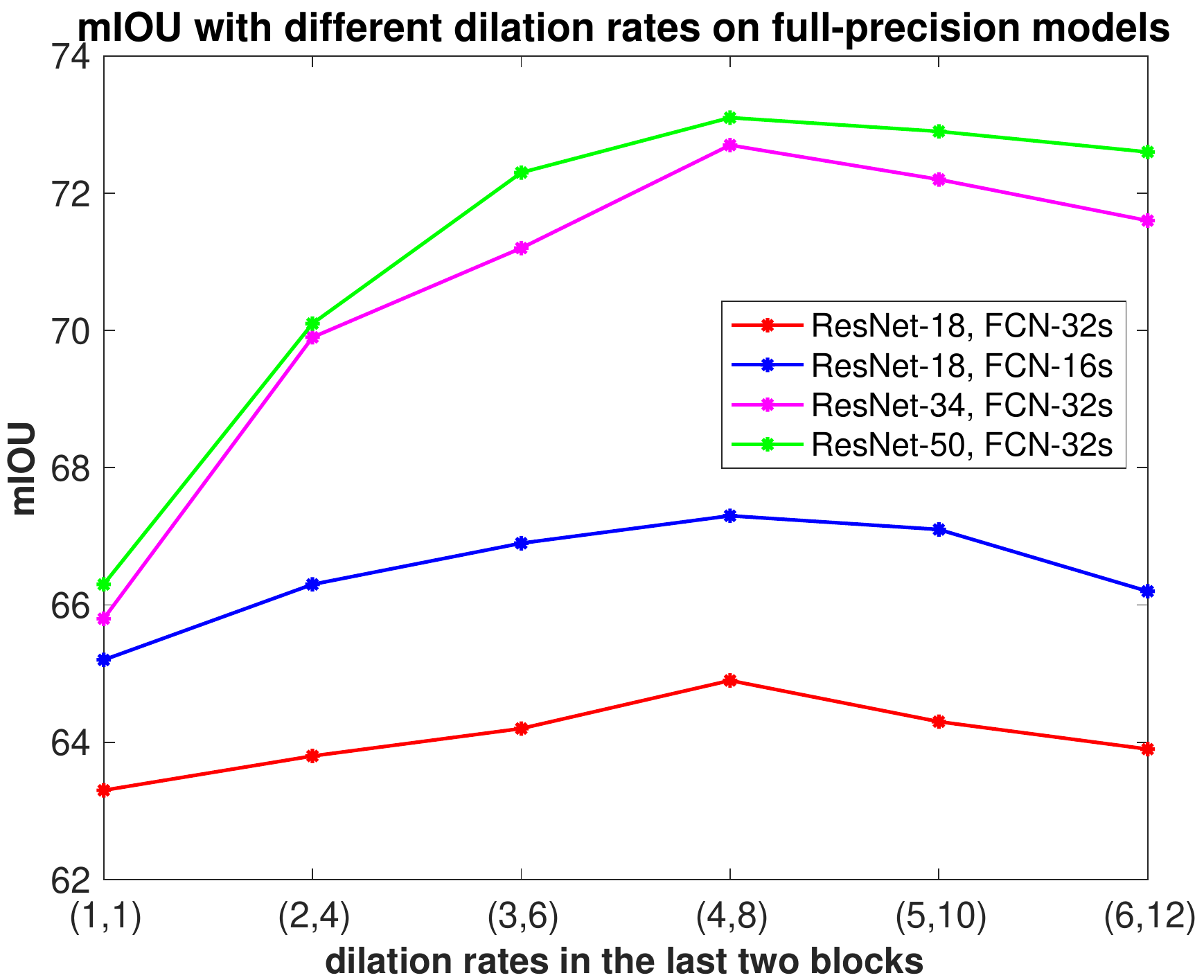}
		\end{tabular}
	}
	\caption{Illustration of the influence of different dilation rates in the last two blocks for the floating-point baseline models.}
	\vspace{-1.0em}
	\label{fig:segmentation_ablation}
\end{figure}

\subsection{Influence of dilation rates on full-precision baselines}
In this section, we explore the effect of dilation rates in the last two blocks for full-precision baselines. We show the mIOU change in Figure.~\ref{fig:segmentation_ablation}. For dilation rates (1, 1), it corresponds to the original FCN baseline~\cite{long2015fully} with no atrous convolution applied. For both FCN-32s and FCN-16s, we can observe that when using dilated convolution with $\emph{rate}=4$ and $\emph{rate}=8$ in the last two blocks respectively, we can get the best performance.

\subsection{Full-precision baselines with multiscale dilations}

\begin{table}[h]
	\centering
	\scalebox{0.75}
	{
		\begin{tabular}{c c |c }
			\hline
			\multicolumn{2}{c|}{Model} &mIOU \\
			\hline
			\multirow{3}{*}{ResNet-18, FCN-32s}
			& Full-precision (multi-dilations) &67.6  \\
			& Full-precision &64.9  \\
			& Group-Net** + BPAC &65.1  \\\hline

			\multirow{3}{*}{ResNet-18, FCN-16s}
		    & Full-precision (multi-dilations) &70.1  \\
			& Full-precision &67.3  \\
			& Group-Net** + BPAC &67.7 \\\hline

			\multirow{3}{*}{ResNet-34, FCN-32s}
			& Full-precision (multi-dilations) &75.0 \\
			& Full-precision &72.7  \\
            & Group-Net** + BPAC &72.8  \\\hline

		    \multirow{3}{*}{ResNet-50, FCN-32s}
		    &Full-precision (multi-dilations) &75.5 \\
		    &Full-precision &73.1 \\
			&Group-Net + BPAC &70.4\\\hline

	\end{tabular}}
	\caption{Performance on PASCAL VOC 2012 validation set.}
	\vspace{-1.0em}
	\label{tab:segmentation_multi}
\end{table}

In Sec. 5.4 in the paper, we have shown that Group-Net with BPAC can accurately fit the full-precision model while saving considerable computational complexity. To explore the effect of multiscale dilations on full-precision models,
we replace the last two blocks as the same structure of BPAC. Specifically, we use $K$ homogeneous floating-point branches in the last two blocks while each branch is different in dilation rate. We set $K=5$ here.
Because of this modification, the FLOPs for full-precision ResNet-18, ResNet-34 and ResNet-50 increases by 2.79$\times$, 3.14$\times$ and 3.13$\times$, respectively.
As shown in Table~\ref{tab:segmentation_multi}, the multiple dilations design improves the performance of full-precision baselines but at a cost of huge computational complexity increase.
In contrast, Group-Net+BPAC does not increase the computational complexity compared with using Group-Net only.
This proves the flexibility of the proposed Group-Net which can effectively borrow task-specific properties to approximate the original floating-point structure. And this is one of the advantages for employing structured binary decomposition.

\section{Extending Group-Net to binary weights and low-precision activations}

In the main paper and in Sec.~\ref{sec:supply_s1} to  Sec.~\ref{sec:supply_s3}, all the experiments are based on binary weights and binary activations. To make a tradeoff between accuracy and computational complexity, we can add more bases as discussed in Sec.~\ref{sec:bases}. However, we can also increase the bit-width of activations for better accuracy according to actual demand. We conduct experiments on the ImageNet dataset and report the accuracy in Table~\ref{tab:bitwidth}, Table~\ref{tab:alexnet} and Table~\ref{tab:bases}.

\subsection{Fixed-point Activation quantization}
We apply the simple uniform activation quantization in the paper.
As the output of the ReLU function is unbounded, the quantization after ReLU requires a high dynamic range. It will cause large quantization errors especially when the bit-precision is low. To alleviate this problem, similar to~\cite{zhou2016dorefa, hubara2017quantized}, we use a clip function $h(y) = {\rm{clip}}(y,0,\beta )$ to limit the range of activation to $[0,\beta]$, where $\beta$ (not learned) is fixed during training. Then the truncated activation output ${\bf{y}}$ is uniformly quantized to $K$-bits ($K > 1$) and we still use STE to estimate the gradient:
\setlength{\abovedisplayskip}{0.5pt}
\setlength{\belowdisplayskip}{1.5pt}
\begin{equation}
\begin{split}
&\mbox{Forward}: {\bf{\widetilde{y}}} = {\rm{round}}({\bf{y}} \cdot \frac{{{2^K} - 1}}{\beta }) \cdot \frac{\beta }{{{2^K} - 1}}, \\
&\mbox{Backward}: \frac{{\partial \ell}}{{\partial {\bf{y}}}} = \frac{{\partial \ell}}{{\partial {\bf{\widetilde{y}}}}}. \\
\end{split}
\end{equation}
Since the weights are binary, the multiplication in convolution is replaced by fixed-point addition. One can simply replace the uniform quantizer with other non-uniform quantizers for more accurate quantization similar to \cite{zhang2018lq, Cai_2017_CVPR}.

\subsection{Implementation details}
For data preprocessing, it follows the same pipeline as BNNs. We also quantize the weights and activations of all convolutional layers except that the first layer and the last layer are full-precision.
For training ResNet with fixed-point activations, the learning rate starts at 0.05 and is divided by 10 when it gets saturated. We use Nesterov momentum SGD for optimization.
The mini-batch size and weight decay are set to 128 and 0.0001, respectively. The momentum ratio is 0.9. We directly learn from scratch since we empirically observe that fine-tuning does not bring further benefits to the performance. The convolution and element-wise operations are in the order: $Conv \to BN \to ReLU \to Quantize$.

\subsection{Evaluation on ImageNet}

For experiments in Table~\ref{tab:bitwidth} and Table~\ref{tab:alexnet}, we use 5 bases (\ie, $K=5$).
From Table~\ref{tab:bitwidth}, we can observe that with binary weights and fixed-point activations, we can achieve highly accurate results. For example, by also referring to Table 2 in the main paper, we can find the Top-1 accuracy drop for Group-Net on ResNet-50 with tenary and binary activations are 1.5\% and 6.5\%, respectively.
Furthermore, our approach still works well on plain network structures such as AlexNet in Table~\ref{tab:alexnet}.
We also provide the comparison with different number of bases in Table~\ref{tab:bases}.

\begin{table*}[h]
	\centering
   \resizebox{0.8\linewidth}{!}
	{
		\begin{tabular}{c c c c c c c}
			\hline
			Model &W&A&Top-1 \%&Top-5 \%&Top-1 gap \%&Top-5 gap \%\\ \hline
			ResNet-18 Full-precision & 32  &32  &69.7 &89.4 &- &-  \\
			Group-Net &1 &2 &69.6 &89.0 &0.1 &0.4 \\
		    Group-Net & 1 & 32 & 70.4 & 89.8 &-0.7 &-0.4 \\
			GBD v1 & 1 &4  &69.2  &88.5  &0.5  &0.9\\
			GBD v2 &1 &4 &68.3 &87.9 &1.4 &1.5 \\
			GBD v3 &1 &4 &64.5 &85.0 &5.2 &4.4 \\
			LBD &1 &4 &60.1 &82.2 &9.6 &7.2  \\
			\hline
			ResNet-50 Full-precision &32 &32 &76.0 &92.9 &- &- \\
			Group-Net &1 &2 &74.5 &91.5 &1.5 &1.4 \\
			Group-Net &1 &4 &76.0 &92.7 &0.0 &0.2 \\\hline
	\end{tabular}}
	\caption{Validation accuracy of different binary decomposition strategies on ImageNet with different choices of W and A. `W' and `A' refer to the weight and activation bitwidth, respectively.}
	\label{tab:bitwidth}
\end{table*}

\begin{table}[h]
	\centering
   \resizebox{1.0\linewidth}{!}
	{
		\begin{tabular}{c c c c c}
			\hline
			Model& Full-precision &LBD &GBD v1 &Group-Net  \\
			\hline
			Top-1 \% & 57.2  &54.2 &57.3 &\bf{57.8} \\
			Top-5 \% & 80.4  &77.6 &80.1 &\bf{80.9} \\\hline
	\end{tabular}}
	\caption{Accuracy of AlexNet on ImageNet validation set. All cases use binary weights and 2-bit activations.}
	\label{tab:alexnet}
\end{table}

\begin{table*}[h]
	\centering
   \resizebox{.8\linewidth}{!}
	{
		\begin{tabular}{c c c c c c c c c}
			\hline
			Model &Bases &bitW  &bitA &Top-1 \%&Top-5 \%&Top-1 gap \%&Top-5 gap \%\\\hline
			Full-precision &1  &32 &32 &69.7 &89.4 &- &-  \\
			Group-Net &1   &1 &4 &61.5 &83.2 &8.2 &6.2 \\
			Group-Net &3  &1 &4 &68.5 &88.7 &1.2 &0.7 \\
			Group-Net &5  &1 &4 &70.1 &89.5  &-0.4 &-0.1\\\hline
	\end{tabular}}
	\caption{Validation accuracy of Group-Net on ImageNet with number of bases. All cases are based on the ResNet-18 network with binary weights and 4-bit activations.}
	\label{tab:bases}
\end{table*}

\clearpage

{\small
\bibliographystyle{ieee}
\bibliography{egbib}
}

\end{document}